\documentclass[letterpaper, 10 pt, conference]{ieeeconf}
\IEEEoverridecommandlockouts
\usepackage{amsmath,amssymb,amsfonts}
\usepackage{graphicx}
\usepackage{booktabs}
\usepackage{multirow}
\usepackage{xcolor}
\usepackage{url}
\usepackage{cite}
\usepackage{balance}

\renewcommand{\footnoterule}{\kern-3pt\hrule width 0.5\columnwidth height 0.4pt\kern 2.6pt}

\begin{document}

\title{\LARGE \bf Beyond Patch Removal: Persistent Adversarial Effects\\ in Vision-Language-Action Policies}

\author{Enhao Wu$^{1,*}$, Fusen Guo$^{2,*}$, Yuxin Cao$^{3}$, Ziyang Lyu$^{1}$, Lin Li$^{4}$, Wei Song$^{5,\dagger}$\\[2pt]
{\normalsize $^{1}$UNSW Sydney, Australia \quad $^{2}$UNSW Canberra, Australia \quad $^{3}$National University of Singapore, Singapore}\\
{\normalsize $^{4}$Southern Cross University, Australia \quad $^{5}$Griffith University, Australia}%
\thanks{$^{*}$These authors contributed equally.}%
\thanks{$^{\dagger}$Corresponding author.}%
}

\maketitle

\begin{abstract}
Adversarial patches to Vision-Language-Action (VLA) policies can cause both immediate action corruption and persistent state effects that remain after the patch is removed. Existing evaluations largely focus on continuous attacks and do not separate these two effects. We introduce a state-restoration protocol that removes the patch at matched action-chunk boundaries and measures subsequent recoverability under the same remaining step budget. Clean, random-patch, deviation-matched, and fixed-direction controls distinguish adversarial effects from occlusion, action-error magnitude, and directional persistence. We also evaluate a recovery adapter trained on attack-induced states under controlled intervention latency. On OpenVLA-OFT with EDPA attacks, only 36.2\% of LIBERO-Long episodes remain recoverable after five chunks, compared with 89.9\% and 87.0\% for the deviation-matched and fixed-direction controls. Similar persistent effects are observed on autoregressive OpenVLA. The recovery adapter improves recovery from 7.7\% to 47.4\% at one-chunk latency, but its benefit decreases substantially with delayed intervention. These results show that adversarial effects can persist after patch removal and that timely intervention is critical for recovery.
\end{abstract}

\section{Introduction}

Vision-Language-Action (VLA) policies map visual observations and language instructions directly to robot actions \cite{brohan2023rt2,kim2024openvla,black2025pi0}. This closed-loop setting makes adversarial perturbations fundamentally different from those in static perception tasks~\cite{brown2017patch,eykholt2018robust,cao2025towards}: an incorrect action changes the physical state of the environment and therefore affects subsequent observations and decisions. However, existing evaluations of adversarial attacks on VLA policies largely focus on task success under a continuously applied patch or on per-step action deviation \cite{wang2025exploring,xu2025edpa}. These metrics characterize the policy while the perturbation is present, but do not measure what happens after it is removed.

This distinction is important because removing a visual perturbation does not restore the robot to the state it would have reached under clean execution. During an attack, corrupted actions can move the robot or surrounding objects into different physical configurations, and the policy must continue from the resulting altered states even after clean observations are restored. An adversarial patch can therefore have two effects: immediate action corruption while the patch is present, and a persistent reduction in recoverability after its removal. The latter is particularly relevant to runtime defenses, since detecting or removing a perturbation may not be sufficient if intervention occurs only after the policy has entered a state with low recoverability.

Recent studies have shown that recoverability decreases as non-adversarial perturbations persist for longer durations \cite{jo2026joint,zhang2026arb4wm}. This establishes that closed-loop errors can accumulate over time, but does not determine whether adversarial optimization introduces an additional effect. In particular, a lower recovery rate after an optimized attack could arise simply from occlusion, larger action errors, or temporally persistent deviations. We therefore ask a more specific question: \emph{does an optimized adversarial patch drive a VLA policy into states that are harder to recover from than suitably matched non-adversarial perturbations?}

\begin{figure*}[t]
\centering
\includegraphics[width=0.95\textwidth]{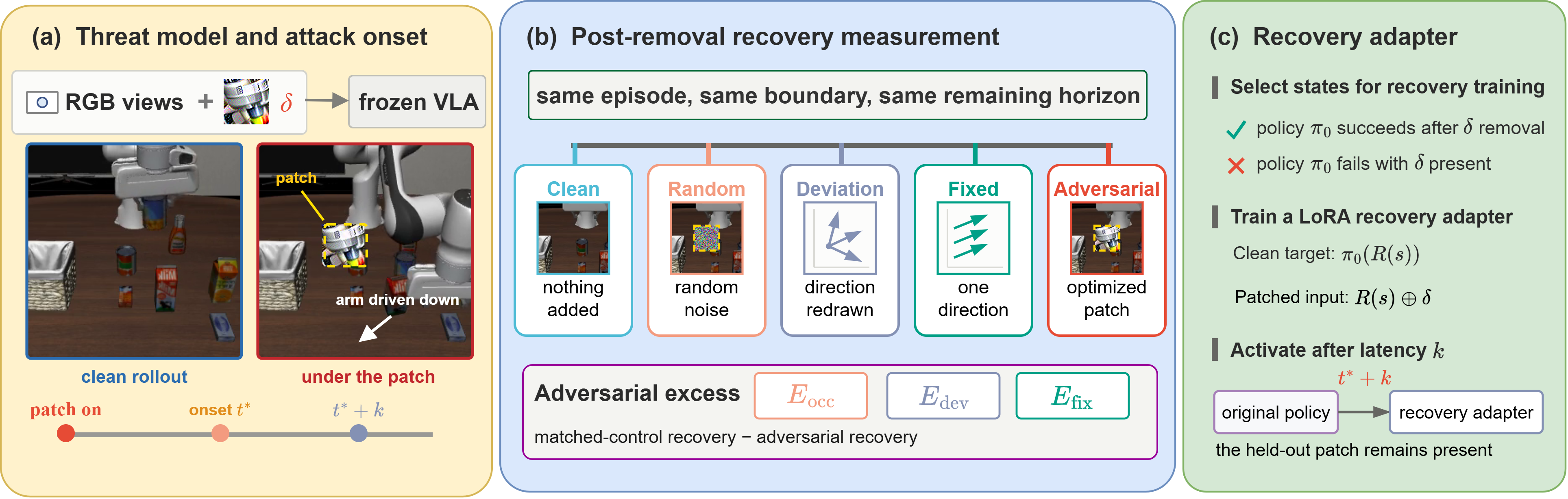}
\caption{Overview of (a) the threat model and attack onset, (b) post-removal recovery measurement with matched controls, and (c) the recovery adapter.}
\label{fig:overview}
\end{figure*}

To answer this question, we introduce a state-restoration protocol for measuring post-removal recoverability, illustrated in Fig.~\ref{fig:overview}. We evaluate frozen OpenVLA-OFT \cite{kim2025oft} policies on LIBERO-Long and LIBERO-Goal \cite{liu2023libero} under the EDPA patch attack \cite{xu2025edpa}. At selected action-chunk boundaries, we restore the simulator to states reached under different perturbation conditions, remove the patch, and resume the same frozen policy with an identical remaining step budget. In addition to clean and random-patch controls, we construct a deviation-matched control that reproduces the adversary's action-error magnitude and a fixed-direction control that additionally reproduces directional persistence. These controls allow us to separate persistent adversarial effects from explanations based on occlusion, perturbation magnitude, or persistent action deviation. We quantify the resulting difference in recoverability as \emph{adversarial excess}.

Our experiments show a substantial persistent effect after patch removal. After five action chunks on LIBERO-Long, only 36.2\% of episodes reached under the adversarial patch remain recoverable, compared with 89.9\% for the deviation-matched control and 87.0\% for the fixed-direction control. The effect is also observed on autoregressive OpenVLA under the occlusion-matched comparison, indicating that it is not specific to chunked action decoding. We further study the recovery implications by training a proof-of-concept recovery adapter on attack-induced states. The adapter improves recovery on LIBERO-Long from 7.7\% to 47.4\% when activated after one action chunk, while substantially less recovery remains available at longer intervention latency.

\noindent\textbf{Contributions.}
Our main contributions are:
\begin{itemize}
    \item We introduce a state-restoration protocol for measuring post-removal recoverability under matched remaining budgets, together with controls for occlusion, action-error magnitude, and directional persistence. The results reveal a persistent adversarial effect that cannot be explained by these factors alone.

    \item We identify and correct three state-restoration fidelity issues in LIBERO that can cause resumed rollouts from the same saved simulator state to diverge, and provide a consistency check for validating exact restoration.

    \item We evaluate a proof-of-concept recovery adapter trained on attack-induced states and show that recovery performance depends strongly on intervention latency, highlighting the need to evaluate VLA defenses jointly in terms of robustness and response time.
\end{itemize}

\section{Related Work}

Vision-Language-Action (VLA) policies map visual observations and language instructions directly to robot actions \cite{brohan2023rt1,brohan2023rt2,oxe2024}. Recent work has examined their vulnerability to adversarial perturbations and robustness under visual and physical variations \cite{wang2025exploring,jones2025attacks,wang2026partial,fu2026hijack,li2025attackvla,fei2025liberoplus,liu2025evavla}. Most adversarial evaluations measure task success under a perturbation that remains present throughout the episode, or quantify action deviations while the policy is under attack. Related defenses focus on preventing or reducing the effect of adversarial patches through certification, localization, purification, or adversarial training \cite{brown2017patch,madry2018towards,xiang2022patchcleanser,tarchoun2023jedi,song2024correction,kang2024diffender,cao2024logostylefool,cao2025towards}. These settings characterize robustness during an attack, whereas we study whether its effect persists after the visual perturbation is removed.

The closest studies to ours explicitly evaluate recovery after a perturbation ends. Jo et al.\ \cite{jo2026joint} inject joint-level physical faults into a VLA policy for different durations and evaluate subsequent performance under fault-free dynamics, showing that recovery decreases as the fault persists. Zhang et al.\ \cite{zhang2026arb4wm} introduce a half-sequence attack setting for world-model agents to evaluate recovery after attack removal. These studies establish that longer perturbations can reduce subsequent task performance. However, they do not compare adversarial trajectories with matched non-adversarial controls, leaving open whether the observed loss of recoverability is caused by adversarial optimization or by perturbation duration and magnitude more generally. Our work addresses this distinction through matched controls for occlusion, action-error magnitude, and directional persistence.

Persistent effects in closed-loop control are also related to error accumulation in imitation learning. A policy can enter states that are poorly represented in its training distribution, causing prediction errors to compound along a trajectory \cite{ross2011dagger,rajaraman2020limits,simchowitz2025pitfalls}. Adversarial attacks on reinforcement-learning policies similarly exploit sequential interactions between actions and future observations \cite{huang2017adversarial,zhang2020robust}. Related work on reversibility and safe reset considers whether an agent can recover from states reached during execution \cite{eysenbach2018leave,grinsztajn2021noturning}. Our setting focuses specifically on states induced by adversarial visual perturbations and measures their effect on subsequent recoverability relative to matched controls.

Runtime monitoring and recovery provide a complementary line of work. Existing methods detect policy failures, estimate execution progress or uncertainty, and invoke recovery policies or safety mechanisms when failures are identified \cite{agia2024sentinel,gu2025safe,xu2025faildetect,dai2025racer,thananjeyan2021recovery}. These methods motivate studying not only whether an intervention succeeds, but also when it occurs. We therefore evaluate recovery as a function of intervention latency and use a proof-of-concept recovery adapter, trained with teacher supervision \cite{hinton2015distilling}, to examine how the time of intervention affects the remaining recoverability.

\section{Method}

\subsection{Post-Removal Recovery Measurement}
\label{sec:recovery_measurement}

\paragraph{Threat model.} Let $\pi$ denote a frozen VLA policy and let $R(s)$ denote the visual observation rendered from environment state $s$. The attacker optimizes a universal visual patch $\delta$ offline and applies it at a fixed image location during execution, as illustrated in Fig.~\ref{fig:overview}(a). The attack is limited to the visual input and does not modify the policy, robot state, or environment dynamics. We consider a digital image-space threat model, so physical projection of a printed patch under changing viewpoints is outside the scope of this work.

A continuous attack affects both the actions predicted while the patch is visible and the environment states reached as a consequence of those actions. We isolate the second effect by removing the patch at a selected boundary and evaluating the frozen policy from the resulting state under clean observations. For a state $s$ and a remaining horizon $H$, we define the following state-conditioned recovery function:
\begin{equation}
Q_{\pi}(s,H) = \Pr\!\left( \text{task succeeds within } H \mid s,\pi,\delta=\varnothing \right).
\label{eq:recovery_function}
\end{equation}
For deterministic policy execution and simulator dynamics, $Q_{\pi}(s,H)\in\{0,1\}$ for a fixed episode.

Let $s^{c}_{e,k}$ denote the state reached in episode $e$ at the $k$-th perturbation boundary following the defined attack onset under condition $c\in\mathcal{C}$, where
\begin{equation}
\mathcal{C} = \{ \mathrm{adv}, \mathrm{clean}, \mathrm{rand}, \mathrm{dev}, \mathrm{fix} \}.
\end{equation}
All conditions are evaluated at the same boundary of the same episode and therefore share the same remaining horizon $H_{e,k}$. The post-removal recovery rate under condition $c$ is
\begin{equation}
\mathrm{SR}_{c}(k) = \frac{1}{|\mathcal{E}_{k}|} \sum_{e\in\mathcal{E}_{k}} Q_{\pi}\!\left(s^{c}_{e,k},H_{e,k}\right),
\label{eq:recovery_rate}
\end{equation}
where $\mathcal{E}_{k}$ contains episodes for which all compared branches are available at boundary $k$.

We measure the persistent effect of the adversarial trajectory through the general contrast
\begin{equation}
\Gamma_{c}(k) = \mathrm{SR}_{c}(k) - \mathrm{SR}_{\mathrm{adv}}(k), c\in \{ \mathrm{clean}, \mathrm{rand}, \mathrm{dev}, \mathrm{fix} \}.
\label{eq:general_contrast}
\end{equation}
The clean-reference contrast $G(k)=\Gamma_{\mathrm{clean}}(k)$ is the overall recovery gap. The three matched-control contrasts are denoted as $E_{\mathrm{occ}}(k)=\Gamma_{\mathrm{rand}}(k)$, $E_{\mathrm{dev}}(k)=\Gamma_{\mathrm{dev}}(k)$, $E_{\mathrm{fix}}(k)=\Gamma_{\mathrm{fix}}(k)$. Because the perturbation is absent during every recovery evaluation, and all branches use the same policy and remaining horizon, these contrasts compare the recoverability of states induced before patch removal rather than the direct effect of the patch on subsequent observations.

\paragraph{Matched controls.} The controls are constructed to test alternative explanations for a lower recovery rate after the adversarial trajectory. The random-patch control uses a patch with the same support and location as $\delta$ but random content, providing a reference for visual occlusion.

To control for action-error magnitude, let $\bar{\mathbf a}_{e,t}=\pi(R(s^{\mathrm{adv}}_{e,t}))$ be the action predicted from the clean rendering of an adversarial-branch state and $\mathbf a^{\delta}_{e,t} =\pi(R(s^{\mathrm{adv}}_{e,t})\oplus\delta)$ the prediction under the patch. For an action vector of dimension $D$, we define the mean absolute deviation operator as:
\begin{equation}
m(\mathbf x)=\frac{1}{D}\|\mathbf x\|_{1}, \qquad \rho_{e,t} = m\!\left( \mathbf a^{\delta}_{e,t} - \bar{\mathbf a}_{e,t} \right),
\label{eq:adv_magnitude}
\end{equation}
where $\rho_{e,t}$ is the adversarial action-deviation magnitude recorded for episode $e$ at boundary $t$.

For the deviation-matched control, an independent random direction $\mathbf u_{e,t}$ is sampled at each boundary and normalized to the same magnitude:
\begin{equation}
\boldsymbol{\eta}^{\mathrm{dev}}_{e,t} = \rho_{e,t} \frac{\mathbf u_{e,t}}{m(\mathbf u_{e,t})}.
\label{eq:dev_control}
\end{equation}
For the fixed-direction control, a single direction $\mathbf u_e$ is sampled once per episode and reused across all boundaries:
\begin{equation}
\boldsymbol{\eta}^{\mathrm{fix}}_{e,t} = \rho_{e,t} \frac{\mathbf u_e}{m(\mathbf u_e)}.
\label{eq:fix_control}
\end{equation}
At each boundary, the corresponding perturbation is added to the action predicted from the clean observation of that control branch before applying the environment action bounds. Thus, both controls match the adversarial deviation magnitude boundary by boundary, while the fixed-direction control additionally imposes strong temporal coherence. Together with the random-patch control, they test whether the observed recovery difference can be explained by occlusion, action-error magnitude, or directional persistence alone.

\subsection{Attack Onset and Exact State Restoration}
\label{sec:restoration}

\paragraph{Attack onset.} Attack duration is measured relative to the point at which the patch first begins to measurably affect the policy's action predictions rather than from the first patched frame. At boundary $t$, we measure
\begin{equation}
d_t = m\!\left( \pi(R(s_t)\oplus\delta) - \pi(R(s_t)) \right).
\label{eq:onset_deviation}
\end{equation}
Let $\tau_{\mathrm{rand}}$ be the $95$th percentile of the corresponding deviation obtained with the matched random patch. We define the attack onset as
\begin{equation}
t^{\ast} = \min \left\{ t: d_t>\tau_{\mathrm{rand}} \;\land\; d_{t+1}>\tau_{\mathrm{rand}} \right\}.
\label{eq:onset}
\end{equation}
Requiring two consecutive threshold crossings reduces sensitivity to isolated action deviations, while calibration against the matched random patch prevents ordinary visual occlusion from spuriously defining the attack onset.

\paragraph{Restorable state.} Post-removal evaluation requires different conditions to be resumed from states recorded at exactly the same execution boundary. A standard simulator snapshot is insufficient for this purpose because the future policy trajectory can also depend on controller history, observation timing, and simulator variables that are not included in the default serialized state.

We therefore represent a resumable checkpoint as
\begin{equation}
z_t = \left( x_t,\, h_t,\, v_t,\, o_t \right),
\label{eq:extended_state}
\end{equation}
where $x_t$ contains the serialized simulator state, $h_t$ contains controller variables required to reproduce subsequent actions, $v_t$ contains task and visualization variables not captured by the standard snapshot, and $o_t$ stores the visual and proprioceptive observation associated with the resume boundary. The cached observation is used for the first policy query immediately after restoration, after which subsequent observations are generated normally by the simulator.

We validate restoration through replay consistency. Given a checkpoint $z_t$ taken from a source rollout, restoring $z_t$ and continuing with the same policy, inputs, and random state must reproduce the original continuation:
\begin{equation}
\hat{\mathbf a}_{t:T} = \mathbf a_{t:T}, \qquad \hat{T}=T, \qquad \hat{y}=y,
\label{eq:replay_consistency}
\end{equation}
where $\mathbf a_{t:T}$ and $\hat{\mathbf a}_{t:T}$ are the original and replayed action sequences, $T$ is the termination step, and $y$ is the task outcome. Only checkpoints satisfying this consistency requirement are used for matched recovery evaluation.

\subsection{Recovery Adapter}
\label{sec:recovery_adapter}

We further use a proof-of-concept recovery adapter to study how intervention latency affects recovery from attack-induced states encountered during execution. Let $\pi_0$ denote the frozen original policy. Candidate states are collected at $k\in\{1,2,5\}$ boundaries after $t^{\ast}$. For a candidate state $s$, define $Y^{-}_{\pi_0}(s)$ as the outcome when $\pi_0$ continues from $s$ after the patch is removed and $Y^{+}_{\pi_0}(s,\delta)$ as the outcome when the patch remains present. We retain
\begin{equation}
\mathcal{S}_{\mathrm{rec}} = \left\{ s: Y^{-}_{\pi_0}(s)=1 \;\land\; Y^{+}_{\pi_0}(s,\delta)=0 \right\}.
\label{eq:recovery_states}
\end{equation}
The first condition ensures that successful behavior is available from the selected state, while the second restricts training to states for which intervention is required.

For each $s\in\mathcal{S}_{\mathrm{rec}}$, the frozen policy under the clean rendering provides the target action $\pi_0(R(s))$, while the adapted policy observes the patched rendering $R(s)\oplus\delta$. We optimize LoRA parameters $\theta$ using
\begin{equation}
\begin{aligned}
\mathcal{L}_{\mathrm{rec}}(\theta) &= \mathbb{E}_{s\sim\mathcal{S}_{\mathrm{rec}}} \left[ \left\| \pi_{\theta}(R(s)\oplus\delta) - \pi_0(R(s)) \right\|_1 \right], \\ \mathcal{L}(\theta) &= \mathcal{L}_{\mathrm{rec}}(\theta) + \lambda \mathcal{L}_{\mathrm{clean}},
\end{aligned}
\label{eq:recovery_objective}
\end{equation}
where $\mathcal{L}_{\mathrm{clean}}$ replays clean expert examples during training to preserve the policy's nominal behavior.

The adapter directly changes the policy behavior in attack-induced states; it does not detect or remove the patch. To isolate the effect of response time, evaluation uses a controlled activation boundary:
\begin{equation}
\pi^{(k)}_t =
\begin{cases}
\pi_0, & t < t^{\ast}+k,\\
\pi_{\theta}, & t \ge t^{\ast}+k,
\end{cases}
\label{eq:switch_policy}
\end{equation}
while the held-out patch remains visible throughout the continuation. The variable $k$ therefore represents intervention latency independently of any particular detection method.

\section{Experiments}

\subsection{Experimental Setup}

\noindent\textbf{Policies and benchmarks.} Our primary evaluation uses the official OpenVLA-OFT \cite{kim2025oft} checkpoints on LIBERO-Long and LIBERO-Goal \cite{liu2023libero}, each containing 10 manipulation tasks implemented in robosuite \cite{zhu2020robosuite}. OpenVLA-OFT predicts eight-step action chunks at 20\,Hz, so five action chunks correspond to two seconds of closed-loop control. To test whether the observed effect depends on chunked continuous-action decoding, we additionally evaluate base OpenVLA \cite{kim2024openvla}, which shares the backbone but autoregressively predicts discrete actions. For this model, states are recorded on the same eight-step temporal grid, and perturbation duration is matched according to elapsed control time.

\noindent\textbf{Attack configuration.} We use EDPA \cite{xu2025edpa} with a $50\times50$ pixel patch, corresponding to the 5\% area ratio used in its reference implementation, and apply the patch to both the third-person and wrist-camera views. Three patches are optimized for recovery training, while three additional patches are optimized using disjoint trajectories and different random seeds for held-out evaluation. The random-patch control uses the same patch size and anchor as the adversarial patch and differs only in pixel content. Evaluation uses the benchmark-provided initial states, whereas recovery-training states are generated from separately sampled initializations, with no initial state shared between training and evaluation.

\noindent\textbf{State-restoration fidelity.} The post-removal evaluation requires a resumed execution to reproduce the source rollout exactly before any experimental condition is changed. We found that a standard LIBERO simulator snapshot omits three quantities required for this property: the Panda gripper's internal controller command, the observation associated with the resume boundary, and task or visualization variables that determine fixture placement and rendering. Omitting the controller state can change the first resumed normalized action by 0.148, while omitted fixture variables can produce pixel differences of up to 198/255 despite an identical serialized physics state. We therefore restore the extended checkpoint defined in Sec.~\ref{sec:restoration} and use the cached camera and proprioceptive observation for the first resumed policy query. With these components restored, replay reproduces the original action sequence exactly, terminates at the same environment step, and yields the same task outcome.

\noindent\textbf{Implementation fidelity.} OpenVLA-OFT relies on bidirectional attention provided by the patched \texttt{transformers} implementation released with the model; the package version string alone does not distinguish this build from the stock release. We use the required patched implementation throughout. We also apply the center crop prescribed for the released LIBERO checkpoints. Both choices were validated by reproducing the published clean-policy performance before running adversarial experiments.

\noindent\textbf{Recovery training and baselines.} We compare four training strategies under a matched optimization budget. Image augmentation applies photometric corruption to clean expert states. Offline adversarial fine-tuning composites the training patches onto clean expert observations at random anchors. A single-duration recovery ablation uses attack-induced states collected only at the shortest duration, while our recovery adapter uses states from $k\in\{1,2,5\}$ with the clean-policy targets defined in Sec.~\ref{sec:recovery_adapter}. The state-selection rule retains 151 of 303 candidate states on LIBERO-Long and 198 of 348 on LIBERO-Goal, yielding 3737 and 2538 supervised frames, respectively. We additionally verify the reuse of recorded attacked-rollout outcomes on 24 boundaries spanning eight tasks and all three collection durations, with all 24 replayed outcomes matching their recorded values. All trained methods use the same base checkpoint, number of training frames, number of optimization steps, and three training seeds. LoRA with rank 32 is applied to every linear layer of the backbone, while the action head and proprioceptive projector remain frozen. Training uses AdamW with a learning rate of $5\times10^{-5}$, cosine decay, and gradient clipping at 1.0. We train for a fixed number of steps and evaluate the final checkpoint without validation-based checkpoint selection.

\begin{figure*}[t]
\centering
\includegraphics[width=0.85\textwidth]{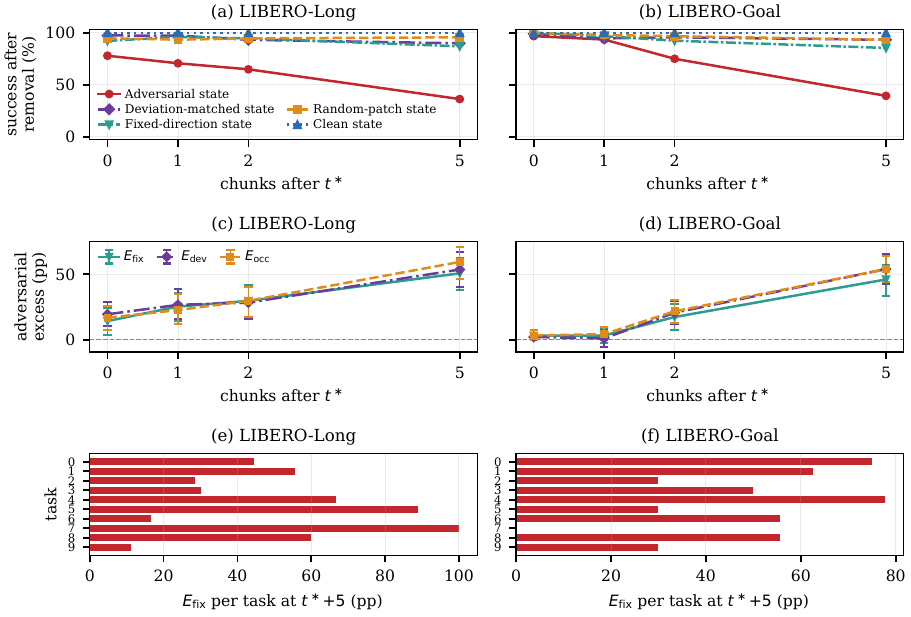}
\caption{Post-removal recovery and adversarial excess as a function of perturbation duration. (a) and (b) show recovery rates on LIBERO-Long and LIBERO-Goal, respectively. (c) and (d) report adversarial excess under the three matched controls. (e) and (f) show task-level $E_{\mathrm{fix}}$ at $t^\ast{+}5$.}
\label{fig:main}
\end{figure*}

\begin{table*}[t]
\caption{Post-removal recovery rates and matched-control adversarial excess.}
\label{tab:depth}
\centering
\small
\begin{tabular}{@{}lccccccccc@{}}
\toprule
& \multicolumn{5}{c}{Recovery rate} & \multicolumn{3}{c}{Adversarial excess (pp)} & \\
\cmidrule(lr){2-6}\cmidrule(lr){7-9}
Duration & Adversarial & Fixed-direction & Deviation-matched & Random & Clean & $E_{\mathrm{fix}}$ & $E_{\mathrm{dev}}$ & $E_{\mathrm{occ}}$ & $n$ \\
\midrule
\multicolumn{10}{@{}l}{\textit{LIBERO-Long}}\\
$t^\ast$ & 77.9\% & 92.2\% & 97.4\% & 94.8\% & 100.0\% & 14.3 & 19.5 & 16.9 & 77 \\
$t^\ast{+}1$ & 70.7\% & 96.0\% & 97.3\% & 93.3\% & 100.0\% & 25.3 & 26.7 & 22.7 & 75 \\
$t^\ast{+}2$ & 64.9\% & 94.6\% & 93.2\% & 94.6\% & 100.0\% & 29.7 & 28.4 & 29.7 & 74 \\
$t^\ast{+}5$ & 36.2\% & 87.0\% & 89.9\% & 95.7\% & 100.0\% & 50.7 & 53.6 & 59.4 & 69 \\
\midrule
\multicolumn{10}{@{}l}{\textit{LIBERO-Goal}}\\
$t^\ast$ & 96.8\% & 100.0\% & 98.9\% & 100.0\% & 100.0\% & 3.2 & 2.1 & 3.2 & 93 \\
$t^\ast{+}1$ & 93.5\% & 96.7\% & 94.6\% & 97.8\% & 100.0\% & 3.3 & 1.1 & 4.3 & 92 \\
$t^\ast{+}2$ & 75.0\% & 92.4\% & 95.7\% & 96.7\% & 100.0\% & 17.4 & 20.6 & 21.7 & 92 \\
$t^\ast{+}5$ & 39.3\% & 85.4\% & 93.3\% & 93.3\% & 100.0\% & 46.1 & 53.9 & 53.9 & 89 \\
\bottomrule
\end{tabular}
\end{table*}

\noindent\textbf{Evaluation and statistics.} The primary recovery measurement uses 100 matched episodes per suite. The recovery-method comparison evaluates three initial states per task at each activation latency. All comparisons are paired by task, initial state, random seed, and patch anchor. We compute 95\% confidence intervals (CIs) using paired bootstrap resampling stratified by task. An episode contributes at duration $k$ only if every compared branch has a valid state at that boundary; consequently, the paired sample size decreases from 77 to 69 on LIBERO-Long and from 93 to 89 on LIBERO-Goal between $t^\ast$ and $t^\ast{+}5$. Because adversarial excess is a difference between success rates, we report it and its confidence interval in percentage points (pp).

\subsection{Persistent Adversarial Effects After Patch Removal}

\noindent\textbf{Continuous-attack baseline.} We first report the conventional evaluation in which the perturbation remains present throughout the episode. Under this setting, EDPA reduces task success from 91.0\% to 9.0\% on LIBERO-Long and from 94.0\% to 20.0\% on LIBERO-Goal. Relative to the matched clean episodes, these correspond to decreases of 82.0 pp and 74.0 pp, respectively. The former has a 95\% confidence interval (CI) of 74.0--90.0 pp. In comparison, the random patch achieves 80.0\% success on LIBERO-Long and 95.0\% on LIBERO-Goal, corresponding to clean-minus-random differences of 11.0 pp and -1.0 pp. These results establish the effectiveness of the adversarial patch under continuous attack, but do not characterize recoverability after the patch is removed.

\noindent\textbf{Post-removal recoverability.} Fig.~\ref{fig:main} and Table~\ref{tab:depth} report recovery after the patch is removed. At every duration, all branches are resumed from the same episode boundary with the same remaining step budget. The clean branch retains 100\% recovery across the paired episodes, indicating that the decrease observed in the adversarial branch is not caused by a shorter remaining horizon. On LIBERO-Long at $t^\ast{+}5$, adversarially induced states recover in only 36.2\% of episodes, compared with 87.0\% for the fixed-direction control, 89.9\% for the deviation-matched control, and 95.7\% for the random-patch control. This yields $E_{\mathrm{fix}}=50.7$ pp, $E_{\mathrm{dev}}=53.6$ pp, and $E_{\mathrm{occ}}=59.4$ pp. For the fixed-direction comparison, the 95\% CI of $E_{\mathrm{fix}}$ is 37.7--62.3 pp. LIBERO-Goal shows the same pattern at $t^\ast{+}5$, with adversarial recovery of 39.3\% and $E_{\mathrm{fix}}=46.1$ pp. The positive excess after matching action-error magnitude and imposing persistent direction indicates that neither factor alone accounts for the reduced recoverability.

\noindent\textbf{Effect of perturbation duration.} The excess increases with duration on both suites, although the early behavior differs. On LIBERO-Goal, all three excess measures remain small at $t^\ast$ and $t^\ast{+}1$, whereas LIBERO-Long already shows a measurable separation at onset. This difference follows from the definition of $t^\ast$: it marks the first sustained action deviation rather than a fixed stage of physical task execution. The median onset on LIBERO-Goal occurs at the first evaluated boundary, environment step 10, before substantial interaction with the scene, whereas the median onset on LIBERO-Long occurs at step 58. The later growth of the excess on both suites indicates that the post-removal effect becomes more pronounced as adversarially influenced actions accumulate.

\noindent\textbf{Task-level consistency.} Fig.~\ref{fig:main} (e) and (f) report $E_{\mathrm{fix}}$ separately for each task at $t^\ast{+}5$. The excess remains positive on all 10 LIBERO-Long tasks and on 9 of 10 LIBERO-Goal tasks; the remaining LIBERO-Goal task has an excess of exactly zero. The aggregate result is therefore not driven by only a small subset of tasks.

\subsection{Generality and Mechanism}

\noindent\textbf{Autoregressive OpenVLA.} We repeat the occlusion-referenced measurement on base OpenVLA, whose action decoder differs from the chunked continuous-action head used by OpenVLA-OFT. Base OpenVLA achieves 54.2\% clean success on LIBERO-Long; continuous EDPA reduces this to 0.8\%, while the matched random patch achieves 24.2\%. Table~\ref{tab:second} reports post-removal recovery on the paired episodes. $E_{\mathrm{occ}}$ increases from 26.5 pp at $t^\ast$ to 39.4 pp at $t^\ast{+}5$, with a 95\% CI of 24.2--57.6 pp at the latter duration. The random patch itself has a larger effect on base OpenVLA than on OpenVLA-OFT, making the occlusion-matched comparison more stringent for this model. Nevertheless, a positive adversarial excess remains after five chunks. This experiment extends the observation to a different action-decoding mechanism, although the two policies still share the same model backbone.

\begin{table}[t]
\caption{Post-removal recovery on base OpenVLA.}
\label{tab:second}
\centering
\small
\begin{tabular}{@{}lccccc@{}}
\toprule
Duration & Adversarial & Random & Clean & $E_{\mathrm{occ}}$ (pp) & $n$ \\
\midrule
$t^\ast$ & 29.4\% & 55.9\% & 100.0\% & 26.5 & 34 \\
$t^\ast{+}5$ & 3.0\% & 42.4\% & 100.0\% & 39.4 & 33 \\
\bottomrule
\end{tabular}
\end{table}

\noindent\textbf{Failure characteristics.} We next examine whether the lower recovery rate reflects irreversible changes to the environment or states that are less favorable for completing the task within the remaining horizon. No object leaves the workspace in any evaluated branch, and all observed failures terminate by exhausting the step budget rather than by simulator failure. On LIBERO-Long, the adversarial branch spends an average of 8.9 steps in contact with the scene, compared with 3.9 for the random-patch branch and 0.5 for the clean branch. In contrast, contact remains negligible across conditions on LIBERO-Goal. These observations indicate that the lower recovery rate primarily reflects reduced ability to complete the task within the available remaining horizon rather than irreversible changes to the scene.

\noindent\textbf{Temporal structure of action deviations.} The matched controls show that action-error magnitude and generic directional persistence are insufficient to explain the recovery gap. We measure the cosine similarity between deviation vectors at consecutive boundaries. On LIBERO-Long, adjacent adversarial deviations have a mean cosine similarity of 0.813, compared with 0.370 for the random-patch deviations. The corresponding values on LIBERO-Goal are 0.699 and 0.241. Thus, adversarial deviations are substantially more temporally coherent than those induced by random occlusion. However, the fixed-direction control imposes persistent direction by construction and remains substantially more recoverable than the adversarial branch. Temporal coherence alone is therefore insufficient to account for the observed recovery gap.

\subsection{Recovery Under Intervention Latency}

\noindent\textbf{Recovery with held-out patches.} Table~\ref{tab:method} evaluates each recovery strategy using held-out patches that were optimized on disjoint trajectories with different random seeds. The adapter is activated at either $t^\ast{+}1$ or $t^\ast{+}5$, while the patch remains visible after activation. Each run contains 26 eligible recovery episodes on LIBERO-Long and 28 on LIBERO-Goal, while clean success is measured over 30 episodes. The held-out patch construction separates evaluation from the patches used to generate recovery-training data.

\begin{table}[t]
\caption{Recovery under held-out patches across two latencies.}
\label{tab:method}
\centering
\small
\resizebox{0.98\columnwidth}{!}{%
\begin{tabular}{lccc}
\toprule
Method & $t^\ast{+}1$ & $t^\ast{+}5$ & Clean \\
\midrule
\multicolumn{4}{@{}l}{\textit{LIBERO-Long}}\\
No defense & 7.7\% & 7.7\% & 93.3\% \\
Image augmentation & $7.7 \pm 0.0$\% & $7.7 \pm 0.0$\% & $91.1 \pm 5.1$\% \\
Offline adversarial FT & $48.7 \pm 2.2$\% & $21.8 \pm 4.4$\% & $90.0 \pm 3.3$\% \\
Recovery (single duration) & $41.0 \pm 4.4$\% & $21.8 \pm 5.9$\% & $92.2 \pm 1.9$\% \\
Recovery (ours) & $47.4 \pm 8.0$\% & $20.5 \pm 2.2$\% & $93.3 \pm 3.3$\% \\
\midrule
\multicolumn{4}{@{}l}{\textit{LIBERO-Goal}}\\
No defense & 21.4\% & 21.4\% & 96.7\% \\
Image augmentation & $21.4 \pm 0.0$\% & $21.4 \pm 0.0$\% & $98.9 \pm 1.9$\% \\
Offline adversarial FT & $85.7 \pm 0.0$\% & $45.2 \pm 2.1$\% & $97.8 \pm 1.9$\% \\
Recovery (single duration) & $84.5 \pm 4.1$\% & $40.5 \pm 2.1$\% & $98.9 \pm 1.9$\% \\
Recovery (ours) & $88.1 \pm 4.1$\% & $39.3 \pm 0.0$\% & $98.9 \pm 1.9$\% \\
\bottomrule
\end{tabular}
}
\end{table}

\noindent\textbf{Effect of adversarial training data.} Image augmentation does not improve recovery over the undefended policy at either tested intervention latency, yielding 7.7\% on LIBERO-Long and 21.4\% on LIBERO-Goal. In contrast, all three methods trained with adversarial observations substantially improve one-chunk recovery: their success rates range from 41.0\% to 48.7\% on LIBERO-Long and from 84.5\% to 88.1\% on LIBERO-Goal. We do not observe a consistent ranking among the three adversarially trained methods across the training seeds and therefore do not claim that the proposed recovery adapter outperforms offline adversarial fine-tuning or the single-duration variant. Clean-task performance remains comparable to that of the undefended policy, indicating that the recovery gains are not accompanied by a clear reduction in nominal success.

\noindent\textbf{Intervention latency.} Delaying activation from one to five chunks substantially reduces recovery for every method trained with adversarial observations. For our adapter, success decreases from 47.4\% to 20.5\% on LIBERO-Long, a reduction of 26.9 pp, and from 88.1\% to 39.3\% on LIBERO-Goal, a reduction of 48.8 pp. At five chunks, the three adversarially trained methods lie within 20.5--21.8\% on LIBERO-Long and 39.3--45.2\% on LIBERO-Goal. Across both suites, the decrease associated with delayed activation is substantially larger than the differences among the adversarially trained strategies at a fixed latency.

\section{Discussion and Limitations}

Our results distinguish robustness during an adversarial perturbation from recoverability after the perturbation is removed. The matched controls show that the persistent recovery gap cannot be explained by visual occlusion, action-error magnitude, or directional persistence alone. Together with the temporal analysis, these results indicate that the effect depends on properties of the attack-induced trajectory beyond these matched factors, although our experiments do not identify the specific mechanism responsible. The recovery experiments also show that the amount of recoverability available to an intervention decreases substantially with latency. These observations suggest that runtime defenses for VLA policies should be evaluated not only by their performance while an attack is present, but also by post-removal recovery and the time at which intervention occurs.

Our study focuses on a single adversarial patch family and two simulated LIBERO suites, so validation across attack types, physical robots, and independent VLA model families remains necessary. The second-policy experiment tests generality across action-decoding mechanisms through the occlusion-referenced comparison, while the recovery adapter is evaluated at predefined activation times and on states from which the frozen policy can recover. The matched controls rule out several simple explanations for the recovery gap, but the temporal and task-dependent structure responsible for the persistent effect remains to be characterized. Extending the evaluation to broader attacks and models, real-world settings, integrated detection and recovery, and deeper mechanism analysis provides several directions for future work.

\section{Conclusion}

We studied the persistent effects of adversarial patches on Vision-Language-Action policies after the visual perturbation is removed. Using exact state restoration and matched controls, we separate post-removal recoverability from the immediate effect of the patch and compare adversarially induced states with controls for occlusion, action-error magnitude, and directional persistence. Recovery decreases substantially with perturbation duration, while the matched controls remain considerably more recoverable, showing that the persistent effect cannot be explained by these factors alone. The same occlusion-referenced effect is observed on autoregressive OpenVLA, and a proof-of-concept recovery adapter further shows that substantially more recovery is available at short intervention latency than after delayed activation. These results motivate evaluating adversarial robustness in VLA policies not only under continuous attack, but also through post-removal recovery and its dependence on intervention latency.

\bibliographystyle{IEEEtran}
\bibliography{refs}

\balance
\end{document}